# Triplet-based Deep Similarity Learning for Person Re-Identification

Wentong Liao[1], Michael Ying Yang[2], Ni Zhan[1], Bodo Rosenhahn[1]
[1]Institute for Information Processing, Leibniz University Hannover
[2]Scene Understanding Group, University of Twente
liao@tnt.uni-hannover.de, michael.yang@utwente.nl

## Abstract

*In recent years, person re-identification (re-id) catches great attention in both computer vision community and industry. In this paper, we propose a new framework for person re-identification with a triplet-based deep similarity learning using convolutional neural networks (CNNs). The network is trained with triplet input: two of them have the same class labels and the other one is different. It aims to learn the deep feature representation, with which the distance within the same class is decreased, while the distance between the different classes is increased as much as possible. Moreover, we trained the model jointly on six different datasets, which differs from common practice - one model is just trained on one dataset and tested also on the same one. However, the enormous number of possible triplet data among the large number of training samples makes the training impossible. To address this challenge, a double-sampling scheme is proposed to generate triplets of images as effective as possible. The proposed framework is evaluated on several benchmark datasets. The experimental results show that, our method is effective for the task of person re-identification and it is comparable or even outperforms the state-of-the-art methods.*

## 1. Introduction

Recently, person re-identification (re-id) catches great attention in both computer vision community and industry because of its potential practical applications, such as surveillance security [31], person tracking in cross-camera scenes, and retrieval of lost children. The goal is to find a query person among a gallery of people images [34, 11]. Influenced by illumination condition, widely varying person poses, resolution, partial occlusion, etc., re-id is still an open challenging and popular task.

Since the milestone work [18], deep learning has great achievements in computer vision for different tasks,such as object recognition [26, 12], semantic segmentation [23, 3], artist style transform [10, 16], and the re-id task [30, 32, 35, 29, 33]. However, it is well known that, a large-scale dataset (e.g. ImageNet [18], which have 1.2 million images with 1000 categories) is the prerequisite for sufficiently training a deep learning model [20]. It often lacks of such large-scale dataset in many specific areas. But many smaller datasets are published by different research groups. Jointly training a deep learning model with all these small datasets is worth trying to alleviate the grate demand of dataset. Furthermore, a dataset which is collected by a research group doesn't vary too much because of limited condition of collecting scene and custom of the collector. For example, the CUHK01 [21] (as shown in Fig. 1(a)) and CUHK03 [22] datasets are captured in a university, where most of the collected person samples are students. The i-LID [36] (Fig. 1(d)) dataset is captured in an airport and the many person are taking luggages. PRID [14] (Fig. 1(c)) is taken from street views, where crosswalks are the main actors. The image resolution in VIPeR [13] changes violently with varying camera views. Combing these data together make the training dataset discrepant a lot, and then the model is trained to learn more general and robust feature representation.

Typical person re-id framework contains two major components: a feature extractor to describe each sample of the dataset and a metric to measure the distance between the the query image and the gallery images. Many existing works research these two components separately and most of them pay more attention on the first one [6, 9]. After extracting the features, a standard distance measure such as L2 distance [8], Bhattacharyya distance (Bhat) [25], etc. is applied to calculate their similarity. Our framework is mainly inspired by [22, 19], which learn features and distance metric jointly by designing a reasonable loss function. Furthermore, different from usual way, more than one image can be feed into the system to learn discriminative feature representations simultaneously.

In this paper, we propose a triplet CNN framework to learn the deep similarity representation, with which the distance within the same person identity is decreased and between different persons is increased. Six datasets are combined together to make the training data vary widely, and then the framework is able to learn more general and robust representation for the task of person re-id. While the framework is feed 3 images: two of them have the same person identity and the 3rd is a different person, there is huge number of such possible combinations in the combined dataset, which makes the training impossible. To address this problem, we introduce a double sampling scheme for training the framework efficiently. An overview of the proposed framework is depicted in Fig. 2. The main technical contributions of this paper are three-fold: First, a double-sampling method is proposed to address the challenge of numerous possible combinations of triplet input for training the proposed deep convolution network without loss of generality. Second, a triplet ranking loss function for making the distance within the same person samples smaller while the distance between the samples of different persons larger. Third, the model is jointly trained on six different datasets.

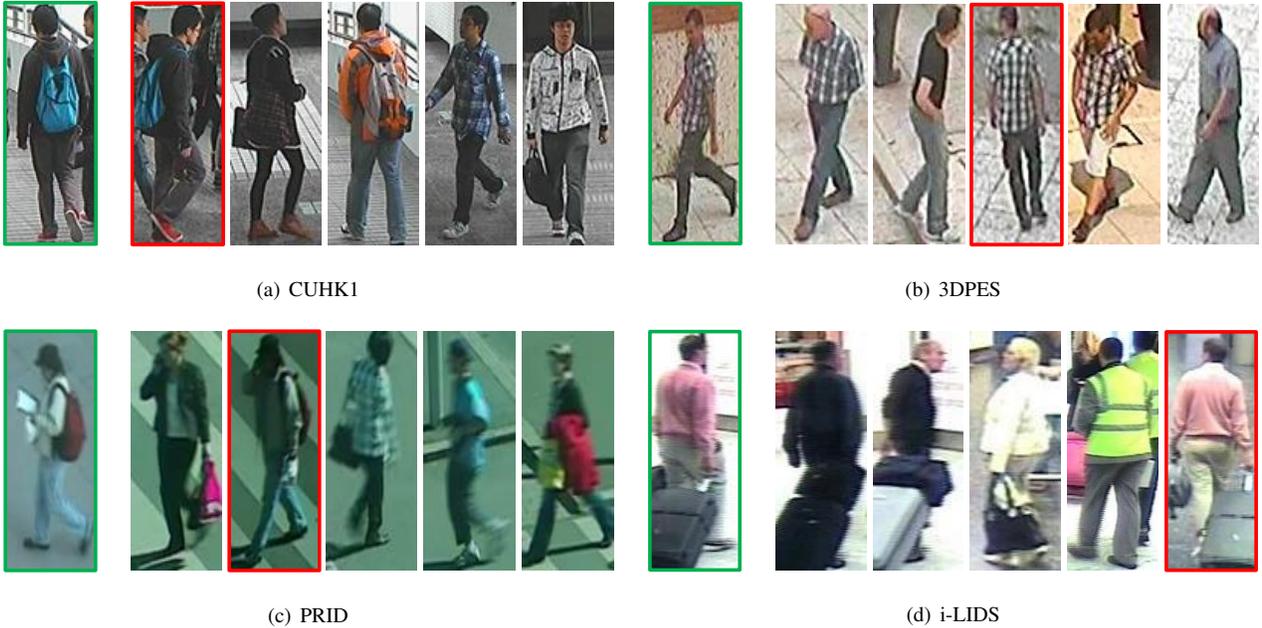

Figure 1: Examples of multiple person re-identification datasets. Each dataset has its certain trait. The green bounding box indicates the query person image and the image in a red box is the corresponding matched person in the gallery.

## 2. Related Work

Before the prevailing of deep learning, manually designing robust and discriminative features to solve computer vision tasks cost most of the researchers' efforts [17, 28, 7]. In recent years, using deep convolutional neural network (CNN) is prevailing in person re-id. Using CNN for person re-id can be roughly divided into three ways. The first way is treating the re-id problem as a classification task [33]. Each person identity is treated as a category class and the framework is trained as a multi-class classifier. For a given person image, the trained framework predicts its identity. However, such methods require sufficient training samples for each person identity. Normally, it performs poorly by predicting the person who has very few samples, which is similar as the large-scale classification task, such as in the Imagenet challenge. The second way is training the network with a pair of images at each time [1, 22]. It learn to predict the similarity which indicates whether the two input images are the same person. The third way can be deemed as an extension of the second one. The network is trained with a triplet inputs [4, 19], as illustrated in Fig. 3. It is trained to minimize the distance within the identity and maximize the distance between different identities. As similar task, the last two methods are also generally adopted for image retrial [19]. Because for person re-id or image retrial, we only care about if two images are matched. In this case, the distance or similarity measure is more practical than classification. Furthermore, it is not limited by the large number of category classes (or identities).

In computer vision and machine learning, the data that belong to a dataset have identical underlying data distribution. Therefore, most of existing models are trained on one dataset each time and also test on the dataset. However, as well known, deep learning methods normally require large-scale data for training to learn deep representations. Most of existing benchmark dataset are not large enough to train a deep learning framework from scratch. Using a trained model which is pretrained on another larger-scale dataset, such as Imagenet, and then fine tuning on the target dataset has been proved to be an effective way and widely applied [20]. Motivated by this, we try to train a framework on multiple different relative small datasets simultaneously. This paper targets on the task of person re-id. Nonetheless, the proposed way, that jointly trains a deep model on multiple different datasets, can be used in other tasks where there are many relative small datasets for the same task.

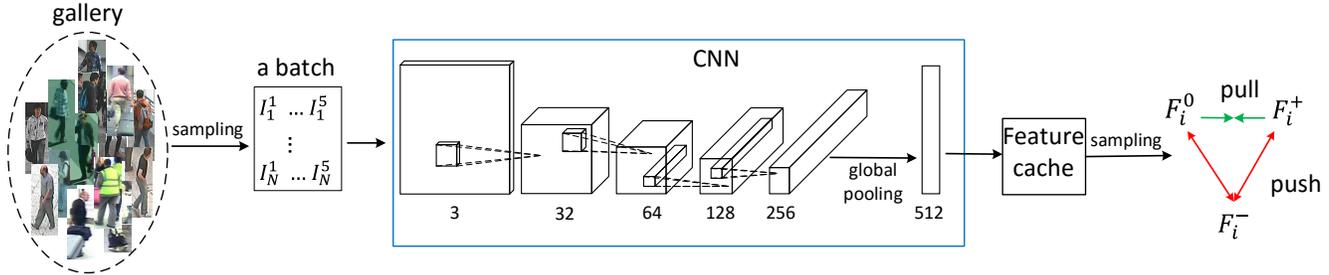

Figure 2: Overview of our framework.

## 3. The Proposed Approach

In this section, we discuss the proposed pipeline for jointly learning deep representations to solve the person re-id problem in details. First of all, we describe the overall framework of our method. Then the proposed triplet structure CNN is presented. Finally, we define the improved triplet-ranking loss function used to train the proposed CNN model.

### 3.1. The Overall Framework

Our proposed pipeline is illustrated in Fig. 2, in which the triplet CNN blocks (as shown in Fig. 3) are merged as a single one because all the CNN blocks share the parameters, i.e., weights and biases. Similar to the works in [19, 4], the proposed network is trained uses triplet examples. The $i-th$ triplet input is denoted as $I_i = (I_i^0, I_i^+, I_i^-)$, where the superscript "0" indicates the anchor image and the "+" means the image of the same person while "−" denotes the image from a different person. A sample image is feed into the CNN model and mapped to the deep feature space $F = f(X)$, where $f(\cdot)$ represents map function of the whole CNN model. $X$ is the input representation of the corresponding image $I$. It's actually the pixel values of the raw image $I$ in this paper. Then the extracted features of the triplet inputs are represented as $F_i = (F_i^0, F_i^+, F_i^-)$. We introduce an improved triplet loss function to calculate the loss for back propagation to guide the neural network to learn a feature space, in which the distance between $F_i^0$ and $F_i^+$ is less than not only the distance between $F_i^0$ (also $F_i^+$) and $F_i^-$. A margin is predefined in the loss to improve the performance as a common way [19, 4].

### 3.2. Double Sampling

Before jointly training the neural networks, we mixed all the benchmark datasets together to get a large-scale gallery of person images. As mentioned above, the CNN is trained with triplet inputs, then a challenging problems we muss overcome is that there can be enormous number of possible combinations of the triplet inputs. For instance, a dataset consists 1000 images from 100 different persons and each person has 10 images, then there are totally $2 * \binom{100}{2} * \binom{10}{2} * \binom{10}{1} = 4455000$ possible combinations of triplet unit. The mixed gallery used in our experiments contains more than 20 thousands images from about 2600 persons. The possible combinations of triplet unit in such dataset is more than 10 billions. On the other hand, if the triplet units are generated first and then fed into the CNN to get the loss, many images go through the CNN repeatedly because a image may be included in more than one triplet units. To handle the combinations of triplet inputs challenge and reduce the unnecessary repeatedly computation, we introduce a double sampling scheme into the pipeline.

The first sampling process happens by generating the mini-batch training data. In the normal ways, all the data are randomly shuffled and then uniformly distributed to each mini-batch. The framework going through all the mini-batches once is a training epoch. Equivalently in the case of triplet training framework, a mini-batch contains a certain number combinations of triplet inputs and a training epoch is feeding all the possible combinations of triplet inputs to the framework. As discussed above, it is costly or impossible. Therefore, a mini-batch is formed in following way. First, 10 different person identities are randomly chosen from the data pool. Then, for each identity 5 raw images are randomly selected. Thus, each mini-batch has 50 images from 5 different persons. The second sampling step is operated before the loss layer to generate the combinations of triple images. The framework extracts features of all the 50 images and saves in cache before the loss layer. One identity is randomly chosen from the 10 in the mini-batch, and then two of the 5 extracted features of this person are randomly chosen from the cache, which are denoted as $F_i^0$, and $F_i^+$ respectively. Another one identity is randomly chosen from the other 9 persons and one of its extracted features is also randomly selected, which is denoted as $F_i^-$. This operation is repeated 2250 times, i.e. 2250 triplet images are generated. 2250 is chosen because it is half of the total number of all possible combinations of triplet images in a mini-batch. The two sampling steps ensure that as many different combinations of triplet images are generated as possible. Furthermore, the second sampling step reduce the repeated computation of same image which is contained in more than one combinations of triplet images. An-other interesting aspect is that, the double sampling method makes the framework be trained to learn discriminative feature representations across different data domains.

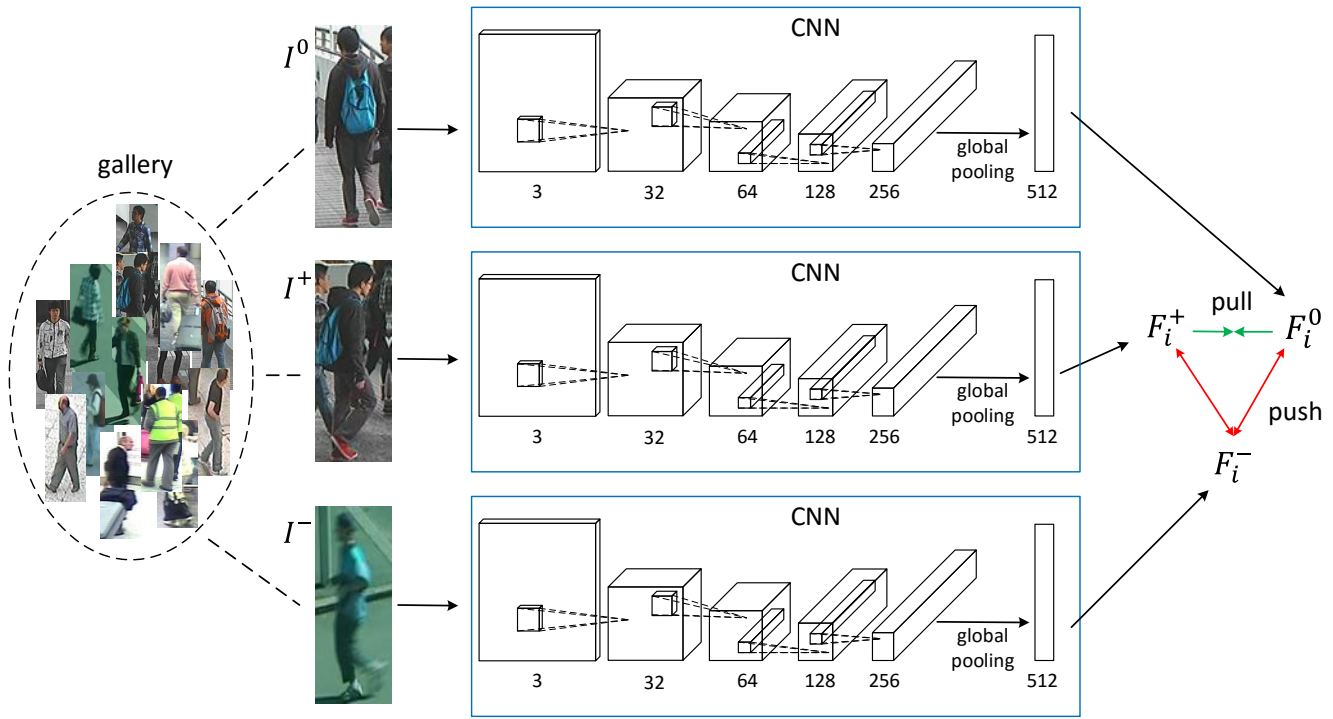

Figure 3: Triplet training framework for person re-id. The three CNN models are parameter shared, i.e. they are the same one. The pairwise input training framework is similar to this way, just delete one CNN block and use a pair of inputs instead the triplet inputs, and the final output is a similarity value to indicates whether the two input images are the same person.

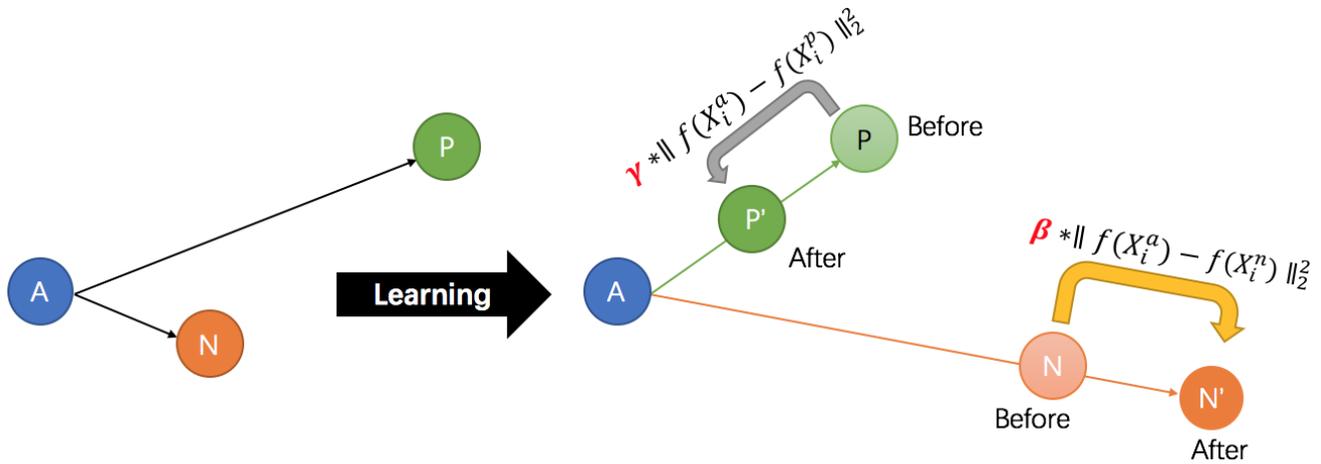

Figure 4: Illustration of the influence of the improved triplet loss function. The letters in the node mean Anchor image, Positive and Negative sample, respectively. The positive sample has the same identity as the anchor while the negative example has a different one. The weights in the improved triplet loss function aims to enlarge the distance between anchor image and the negative sample, and reduce the distance between the samples from the same person.

### 3.3. Improved Triplet Loss Function

The original triplet loss function requires that the distance between the samples of the same person $(F_i^0, F_i^+)$ is smaller than that between different person $(F_i^0, F_i^-)$. It's formally defined as:

$$L = max\{0, \| F_i^0 - F_i^+ \|_2^2 - \| F_i^0 - F_i^- \|_2^2 + \alpha\} \quad (1)$$

where $\| \cdot \|_2^2$ is the squared Euclidean distance, and $\alpha$ is a predefined margin which regularizes the distance, which is normally fixed as 1 and we also adopt in this paper.

The closer the distance of $(F_i^0, F_i^+)$ is and at the same time the further the distance of $(F_i^0, F_i^-)$ is, the more discriminative and robust representations the trained model can learn to describe how much two the two given images are similar. Therefore, two parameters are added to the two distance terms in Eq. (1) respectively, and the equation is rewritten as:

$$L = max\{0, \gamma \| F_i^0 - F_i^+ \|_2^2 - \beta \| F_i^0 - F_i^- \|_2^2 + \alpha\} \quad (2)$$

The impact of three different parameters in Eq. (2) is illustrated in Fig. 4 and will be presented and discussed in details in the experiments.

## 4. Experiments

### 4.1. Datasets

In our experiment, six benchmark datasets are used to train and validate the proposed framework. CUHK03 [21] is one of the most largest widely used dataset in person re-id field. It contains over 14,000 images of 1467 people which are captured from five different pairs of camera views. CUHK01 [22] is collected form the same campus as CUHK03. The difference is that the 1552 images are taken from two camera views. Images in PRID [5] are pedestrians that are cropped from two video sequences which are recorded from different cameras. There 200 people appeared in both views, which are used in our experiments. VIPeR [13] is one of the most challenging dataset for person re-id. Because the images of the 632 people are taken in different poses, from different viewpoints. Moreover, the image resolutions and illumination conditions varies much. 3DPeS [2] records the images from 193 people, and each of them has different number of samples i-LIDS [36] contains 479 images of 119 passengers in the airport. Each person has four images in average. Some images from these datasets are shown in Fig. 1. All the raw images are resized to $144 * 56$ uniformly.

### 4.2. Evaluation

The training and test sets are mainly split following the settings in [24]. Tab. 1 gives a summary of the six datasets for training, validation and test. To quantitatively evaluate the experimental results, the widely used cumulative match curve (CMC) metric is adopted in our experiments. For each query image, we first compute the distance between the query image and all the gallery images using the $L_2$ distance with the features extracted by the trained network, and then return the top $n$ images which have the smallest distance to the query image in the gallery set. If the returned list contains at least one image belonging to the same person as the query image, then this query is considered as success of top $n$. Top $1, 5, 10$ and $20$ are used in this paper. The experiments are repeated 10 times, and the average rate is used as the evaluation result.

The main structure (i.e. the CNN block) of the proposed framework adopts the architecture proposed in [27], as shown in Fig. 5. The difference is all the kernel size of the convolution layers before the first inception block is $3 \times 3$ and the size of $fc7$ is set 512 because of better performance in the experiments. Our framework is implemented in the open source Caffe framework [15]. The initial learning rate is set to 0.01 and is decreased by $5\%$ after every 50 epochs until it reaches 0.0005. In our framework, a training epoch runs only one mini-batch because of the particularity of the double sampling scheme. Therefore, the framework is trained with at least 10 thousand epochs.

### 4.3. Experiments results

The accuracy of person re-id on the six datasets given by the trained framework is listed in Tab. 2. In order to validate the proposed improved triplet loss function, we compare the experimental results with the ones given by our framework but use the normal triplet loss function Eq. (1) which is listed in Tab. 3. We can see that, the framework gains obviously with the improved triplet loss function on all the six datasets.

We explore the influence of the parameters in Eq. (2). Different parameters are tried by grid searching. Some of the experimental results on the largest dataset CUHK03 are listed in Tab. 4. We observe that $(\gamma = 1, \beta = 0.3)$ give the best performance on the Top 1 accuracy, which we care the most for person re-id task. Fig. 6 gives an overview of the influence of the newly added weights on the 6 datasets. We can see that, the best parameters are very similar: $\gamma = 1$ and $\beta$ is around 0.3. It's interesting to find that, the best results are always given by $\gamma = 1$. Because it is equivalently to normalize the each term with $\gamma$ in Eq. (2).

We compare the experimental results output by our method with the ones given by the state-of-the-art [32, 30, 33]. The comparison is given in Tab. 5. From the table we can see that, our method outperforms the methods proposed

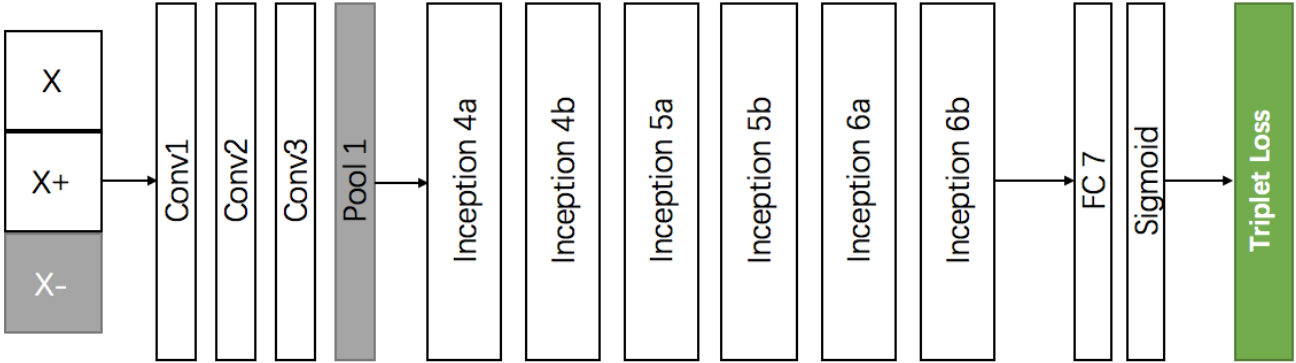

Figure 5: The neural network we adopt in our framework.

| Dataset | # ID | # Trn. images | # Val. images | # Val. ID |
|---|---|---|---|---|
| CUHK03 | 1467 | 21012 | 5252 | 100 |
| CUHKO1 | 971 | 1552 | 3889 | 485 |
| PRID | 385 | 2997 | 749 | 100 |
| VIPeR | 632 | 506 | 126 | 316 |
| 3DPeS | 193 | 420 | 104 | 96 |
| i-LIDS | 119 | 194 | 48 | 60 |

Table 1: Statistics of the datasets for experiments.

|  | Top 1 | Top 5 | Top 10 | Top 20 |
|---|---|---|---|---|
| CUHK03 | 41.0 | 77.7 | 87.3 | 94.3 |
| CUHKO1 | 31.5 | 59.1 | 69.9 | 79.2 |
| PRID | 8.0 | 25.0 | 31.0 | 41.0 |
| VIPeR | 15.2 | 34.2 | 45.9 | 61.1 |
| 3DPeS | 27.2 | 50.8 | 62.9 | 76.7 |
| i-LIDS | 32.1 | 56.9 | 68.7 | 82.9 |

Table 2: Accuracy with improved triplet-loss and double online sample on different datasets.

|  | Top 1 | Top 5 | Top 10 | Top 20 |
|---|---|---|---|---|
| CUHK03 | 38.3 | 73.4 | 85.9 | 94.7 |
| CUHKO1 | 19.8 | 41.6 | 53.2 | 68.0 |
| PRID | 5 | 14 | 23 | 27 |
| VIPeR | 11.4 | 28.5 | 39.2 | 48.1 |
| 3DPeS | 20.9 | 40.6 | 51.8 | 66.5 |
| i-LIDS | 25.9 | 47.4 | 59.5 | 72 |

Table 3: Accuracy with normal triplet-loss and double online sample on different datasets.

| $(\gamma, \beta)$ | Top 1 | Top 5 | Top 10 | Top 20 |
|---|---|---|---|---|
| (10, 0.1) | 39.7 | 76.5 | 87.7 | 94.9 |
| (2, 0.2) | 39.3 | 76.2 | 87.4 | 94.9 |
| (1.2, 0.2) | 41.0 | **76.7** | 87.3 | 94.3 |
| (1, 0.3) | **41.8** | 76.5 | 87.3 | 94.9 |
| **(1, 0.5)** | 40.4 | 76.5 | **87.8** | **95.9** |

Table 4: Experimental results on CUHK03 with different weights.

| CUHK03 | Top 1 | Top 5 | Top 10 | Top 20 |
|---|---|---|---|---|
| **OURS** | 56.1 | **84.4** | **91.0** | **95.2** |
| DeepFeatures [30] | 33.22 | 59.08 | 70.16 | 82.46 |
| Personnet [32] | 40.00 | – | 80.51 | 91.08 |
| JSTL [33] | **72.0** | – | – | – |

Table 5: Comparison of experimental results on the CUHK3 between our method and the state-of-the-art. Some results are not given by the methods, which are indicated by a minus sign.

in [32, 30] in all kinds of accuracy measure significantly. Notice that our model is jointly trained on the six datasets from scratch, while the compared methods used pretrained model (on Imagenet) and fine tuned on the dataset which is used for test. From the comparison we see that our method learns more robust and discriminative deep features from different datasets simultaneously. On the other hand, the method presented in [33] performs better than ours. They also jointly trained their model on the six datasets, but they furthermore used a novel domain dropout methods, which choose the most effective neurons in the training step as the

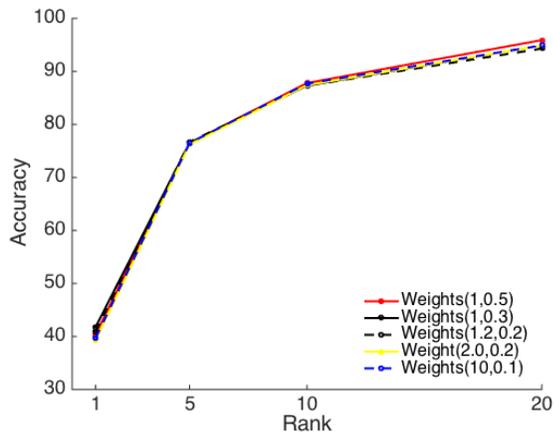

(a) CUHK03

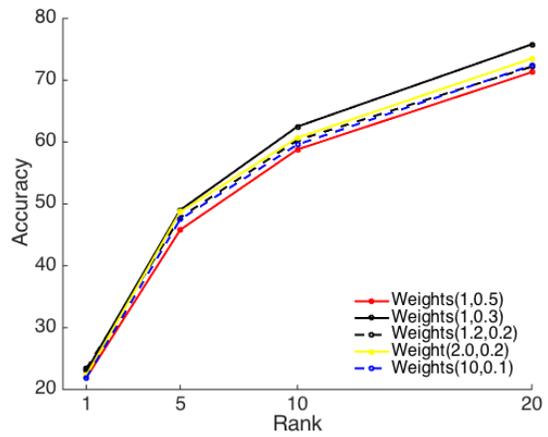

(b) CUHK01

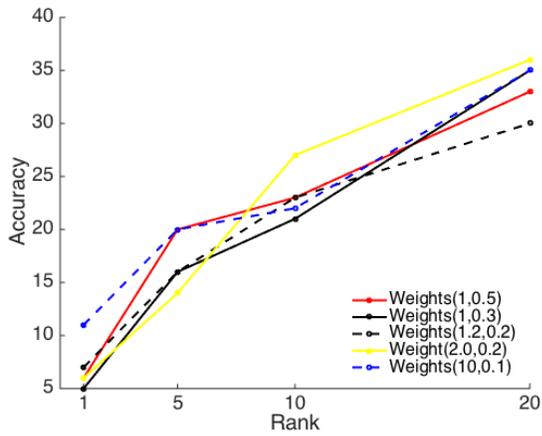

(c) PRID

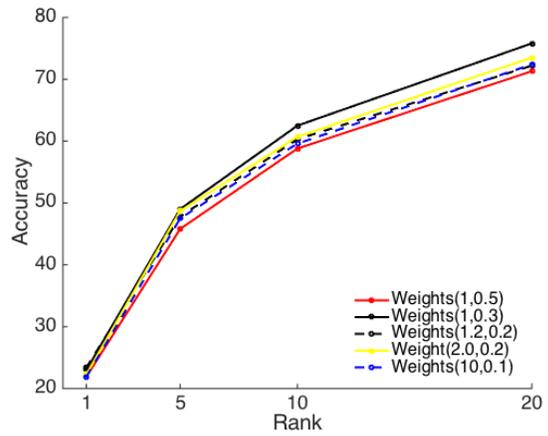

(d) VIPeR

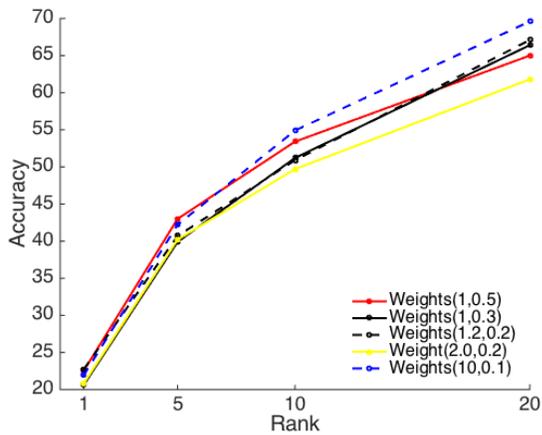

(e) 3DPeS

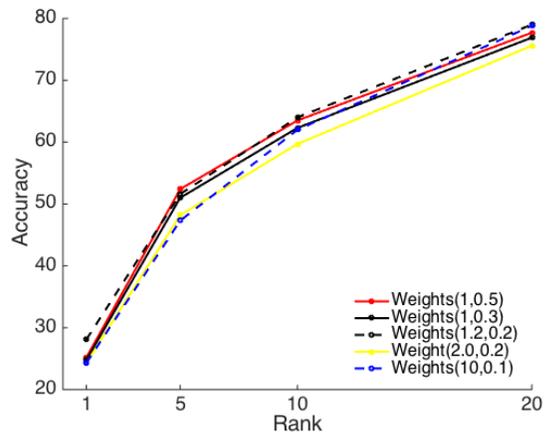

(f) i-LIDS

Figure 6: Influence of different weights on different datasets.

active neurons in the test step for specific dataset. For instance, during the training process, it calculates the impact of each each neuron on a specific dataset, e.g. CUHK3, and then record the most active neurons. When the test image is from CUHK3, only the recorded neurons respond to the input. Their idea inspire us for future work to improve the performance of our framework.

## 5. Conclusions

In this paper, we present a triplet training framework to jointly to learn robust and discriminative deep feature representations on six datasets for person re-id problem. A double sampling method is introduce to overcome the problem of almost infinity possible combinations of triplet inputs on the mixed large-scale dataset. A new triplet loss function is proposed by adding new weights to reduce the distance between the images of the same person and to enlarge the distance between the images of different persons. The proposed framework is evaluated on several benchmark datasets. The experimental results show that, our method is effective for the task of person re-identification and it is comparable or even outperforms the state-of-the-art methods.

## Acknowledgments

The work is funded by DFG (German Research Foundation) YA 351/2-1 and RO 4804/2-1. The authors gratefully acknowledge the support.